\documentclass[preprint]{elsarticle}

\usepackage[english]{babel}

\usepackage[letterpaper,top=2cm,bottom=2cm,left=3cm,right=3cm,marginparwidth=1.75cm]{geometry}

\usepackage{natbib}
\setcitestyle{authoryear,open={(},close={)}} 

\usepackage[utf8]{inputenc} 
\usepackage[T1]{fontenc}    
\usepackage{hyperref}       
\usepackage{url}            
\usepackage{booktabs}       
\usepackage{amsfonts}       
\usepackage{nicefrac}       
\usepackage{microtype}      
\usepackage{xcolor}         
\usepackage{subcaption}

\usepackage{amsmath}
\usepackage{amssymb}
\usepackage{graphicx}
\usepackage{multicol}
\usepackage{multirow}

\usepackage{verbatim}
\usepackage{caption,subcaption}
\usepackage[overload]{empheq}
\usepackage{cleveref} 
\usepackage{amsthm}
\usepackage{epsf, epsfig, hyperref, mathtools,amsthm,bbm,xcolor,pifont}
\usepackage{algorithm}
\usepackage{algpseudocode}
\usepackage{indentfirst}

\numberwithin{equation}{section}

\title{A competitive baseline for deep learning enhanced data assimilation using conditional Gaussian ensemble Kalman filtering}
\author[label1]{Zachariah Malik}
\author[label2]{Romit Maulik}
\affiliation[label1]{organization={Department of Applied Mathematics, University of Colorado Boulder},
         city={Boulder},
         postcode={80309},
         state={CO},
         country={United States of America}}

\affiliation[label2]{organization={College of Information Sciences and Technology, Pennsylvania State University},
         city={University Park},
         postcode={16802},
         state={PA},
         country={United States of America}}

\begin{document}

\begin{abstract}
Ensemble Kalman Filtering (EnKF) is a popular technique for data assimilation, with far ranging applications. However, the vanilla EnKF framework is not well-defined when perturbations are nonlinear. We study two non-linear extensions of the vanilla EnKF - dubbed the conditional-Gaussian EnKF (CG-EnKF) and the normal score EnKF (NS-EnKF) - which sidestep assumptions of linearity by constructing the Kalman gain matrix with the `conditional Gaussian' update formula in place of the traditional one. We then compare these models against a state-of-the-art deep learning based particle filter called the score filter (SF). This model uses an expensive score diffusion model for estimating densities and also requires a strong assumption on the perturbation operator for validity. In our comparison, we find that CG-EnKF and NS-EnKF dramatically outperform SF for a canonical problem in high-dimensional multiscale data assimilation given by the Lorenz-96 system. Our analysis also demonstrates that the CG-EnKF and NS-EnKF can handle highly non-Gaussian additive noise perturbations, with the latter typically outperforming the former.
\end{abstract}

\maketitle

\section{Introduction}
Data assimilation (DA) refers to a class of techniques that combine observations with a model in order to improve predictions of the state of some dynamical system. DA is used in a wide variety of applications, including atmospheric and ocean sciences \citep{Slivinski19,Hu21} , hydrology \citep{Zhou11}, geosciences \citep{Pandya22}. 

One important problem encompassed under DA is the \textit{Bayesian filtering problem}. This refers to the task of finding the posterior on the current state given all past and current observations. One method to solve the Bayesian filtering problem (heretofore referred to as simply the filtering problem) is to perform Kalman filtering. It has been shown that the Kalman filter exactly solves the Bayesian filtering problem \citep{Jazwinski70} when all dynamics and observations are linear and all distributions involved are Gaussian. Due to the computational cost of Kalman filtering, they are not often used in practical applications \citep{Ahmed20}. In response to this, \citet{Evensen94} developed a Monte-Carlo based approximation to the Kalman filter, dubbed the ensemble Kalman filter (EnKF), which has since proven incredibly popular \citep{Evensen03}, \citep{Ahmed20}. 

EnKF is generally expected not to converge to the correct Bayesian posterior in nonlinear, non Gaussian systems \citep{Grooms22} and has engendered a host of extensions as possible remedies to this limitation. The first under consideration in this work is the version of EnKF used in \cite{Grooms22}. Since this version is not the traditional version of EnKF, we refer to it as `conditional Gaussian' EnKF, or CG-EnKF. We name it so because \cite{Grooms22} referred to the specific update formula used in this filter as 'the conditional Gaussian' formula. This formula is different from the classic EnKF update step and, as we shall see, is applicable for nonlinear perturbations. Regardless, both the classic EnKF update and the conditional Gaussian update assume that the observations and priors have a joint Gaussian distribution (from which either the EnKF or conditional Gaussian update formulas may be derived). \cite{Zhou11} proposed the so-called normal-score ensemble Kalman (NS-EnKF) filter to try to relax the Gaussian assumption on the prior. This model uses the normal-score transform to map data from its current, non-atomic (or continuous) distribution to a standard normal distribution. In other words, \citet{Zhou11} transforms the ensemble to a \textit{latent} space that satisfies the Gaussian assumption. Such an idea was combined with the update formula for CG-EnKF in \cite{Grooms22}, which we refer to as NS-EnKF (as opposed to the NS-EnKF described by \cite{Zhou11}).

There has also been a concerted effort to use machine learning (ML) techniques in DA, as covered in the reviews by \citet{Buizza22} and \citet{Cheng22}. Indeed, both reviews discuss combining a neural network with EnKF to iteratively emulate hidden dynamics and predict future states. Nevertheless, ML algorithms are typically far too expensive for practical applications. \citet{Chattopadhyay23} navigates this issue by using a pre-trained model to compute covariance matrices for EnKF which accelerate EnKF significantly. Absent from \citet{Chattopadhyay23} is an attempt to deal with nonlinear, non-Gaussian perturbations to the data.

Another leading method to solve the Bayesian filtering problem is \textit{particle filtering} \citep{Leeuwen19}. This refers to a class of Monte-Carlo methods that use sequential importance sampling and estimate the density of the samples themselves \cite{Leeuwen19}, \cite{Grooms22}. Importantly, particle filters do not assume a linear perturbation of the noise (although they do assume Gaussian additive noise) \cite{Bao24}. Classical particle filters suffer from the `curse of dimensionality', so \cite{Bao24} proposed Score Filter (SF) for greater performance. SF is a particle filter proposed by \cite{Bao24} which combines state of the art generative modelling with deep learning algorithms with the particle filter to solve the filtering problem. From \cite{Bao24}, it is clear that SF outperforms vanilla EnKF in a variety of DA problems.

The goal of this paper is to compare the efficacy of CG-EnKF and NS-EnKF against SF. We stress that NS-EnKF can handle highly non-Gaussian noise perturbations, while SF cannot. This is because NS-EnKF ensures the typical assumptions for EnKF hold by transforming variables into a latent space. We produce a comprehensive study comparing each technique on solving the filtering problem on the Lorenz-96 (L96) system which is a benchmark for validating DA algorithms across disciplines \citep{Bao24,Grooms22, Ahmed20}.

Our paper is structured as follows:
\begin{itemize}
    \item In Section $2$ we provide a brief overview of ensemble Kalman filtering and point out an inconsistency in many Kalman filtering implementations.
    \item In Section $3$ we summarize SF and discuss some potential drawbacks on a theoretical level.
    \item In Section $4$ we provide a comprehensive study on performing DA with the Lorenz-96 system. CG-EnKF clearly outperforms SF on solving these test problems.
\end{itemize}

\section{Ensemble Kalman Filtering}

Consider a model $\mathcal{M} : \mathcal{I} \times \mathbb{R}^{d} \rightarrow \mathbb{R}^{d}$ and let $\mathcal{T}(t) : \mathcal{I} \rightarrow \mathbb{R}^{d}$ be a trajectory produced by $\mathcal{M}$ for some corresponding initial condition. In other words, along some fine discretization $\{ t_{i}^{f} \}$, we have $\mathcal{T}(t_{i+1}^{f}) = \mathcal{M}(\mathcal{T}(t_{i}^{f}))$. For simplicity, we let $\mathcal{I}=[0,\alpha]$ for some $\alpha < \infty$. Suppose we have access to an `observation' operator $\mathcal{N} : \mathbb{R}^{d} \rightarrow \mathbb{R}^{d}$ which allows us to access a perturbed value from the ground truth trajectory. In practice, this may be an experimental device to collect data and can be expensive to use. As such, the observations $\mathbf{y}(t) = \mathcal{N}(\mathcal{T}(t))$ may only be available on a much coarser time grid than that of $\mathcal{T}$. The driving question in data assimilation is: how may we most efficiently and effectively reconstruct $\mathcal{T}$ using measurements $y(t)$ and the model $\mathcal{M}$? 

Ensemble Kalman filtering aims to solve this problem by first assuming that the observation operator takes the form
\begin{equation} \label{Assumption: Linear observations}
    \mathcal{N}(\mathcal{T}(t)) = H \mathcal{T}(t) + \epsilon
\end{equation}
where $\epsilon$ is a Gaussian distributed noise term with mean $0$ and covariance matrix $R$. We shall denote this by $\epsilon \sim N(0,R)$. Suppose we have access to observations (obtained through $\mathcal{N}$) on the coarse time grid $\{ t_{k}^{c} \} \subset \{ t_{i}^{f} \} \subset \mathcal{I}$. We wish to produce an ensemble of trajectories $\{ x_{j}(t) \}_{j=1}^{n} : \mathcal{I} \rightarrow \mathbb{R}^{d}$ which approximate $\mathcal{T}(t)$. To do this, we set some initial condition $x(0)$ and propagate the ensemble through the model until reaching $t_{0}^{c}$. For consistency with \cite{Evensen03}, we shall call this the \textit{forecast} ensemble, and denote it by $x_{j}^{f}(t)$. To be precise, the forecast ensemble is produced by the recursion
\begin{equation}
    x_{j}^{f}(t^{f}_{i+1}) = \mathcal{M}(x_{j}^{f}(t^{f}_{i})), \quad \forall t_{i+1}^{f} \in [0, t_{1}^{c}].
\end{equation}
At time $t_{i}^{f} = t_{1}^{c}$, we perform the Kalman filter update. First, we compute the covariance $P_{e}$ of the ensemble $\{ x_{j}^{f}(t_{1}^{c}) \}$. Next, we need an `observation ensemble' defined by
\begin{equation}
    y_{j}(t^{c}_{1}) = H \mathcal{T}(t_{1}^{c}) + \epsilon_{j}, \quad j=1,\cdots,n,
\end{equation}
where each $\epsilon_{j} \sim \mathcal{N}(0,R)$. With these ingredients, we compute the \textit{analysis} ensemble $\{ x_{j}^{a}(t_{1}^{c}) \}$ by
\begin{equation} \label{Eq: Kalman update}
    x_{j}^{a}(t_{1}^{c}) = x_{j}^{f}(t_{1}^{c}) + P_{e}H^{T}(HP_{e}H^{T}+R)^{-1}(y_{j}(t_{1}^{c}) - Hx_{j}^{f}(t_{1}^{c})).
\end{equation}
This formulation was derived in \cite[Section 3.4.3]{Evensen03} and the interested reader is encouraged to consult that source. To keep in line with \cite{Evensen03}, we will define the \textit{Kalman} gain matrix by
\begin{equation}
    K = P_{e}H^{T}(HP_{e}HT^{T} + R)^{-1}
\end{equation}
and for shorthand, we denote the right hand side of \eqref{Eq: Kalman update} with the operator $\mathcal{F}$. Using this notation, we produce the analysis ensemble by
\begin{equation}
    x^{a}_{j}(t_{i}^{f}) = \begin{cases}
        x^{f}_{j}(t_{i}^{f}), \quad t_{i}^{f} \in [0, t_{1}^{c}) \\
        \mathcal{F}(x_{j}^{f}(t_{i}^{f}), y_{j}(t_{i}^{f})), \quad t_{i}^{f} = t_{1}^{c}.
    \end{cases}
\end{equation}
In order to propagate beyond $i$ such that $t_{i}^{f} = t_{1}^{c}$, we simply push the analysis ensemble through the model, that is
\begin{equation}
    x_{j}^{f}(t_{i+1}^{f}) := \mathcal{M}(x_{j}^{a}(t_{i}^{f})).
\end{equation}
Then, push the forecast ensemble until time point $t_{2}^{c}$ and perform another Kalman filter step. This is repeated until the final time step at time $\alpha$. The complete process is detailed in Algorithm~\ref{Alg: EnKF}

\begin{algorithm}
    \caption{Traditional EnKF} \label{Alg: EnKF}
    \begin{algorithmic} 
        \State Given reference trajectory $\mathcal{T}$, dynamical model $\mathcal{M}$, initial ensemble $\{ x_{j}(0) \}_{j=1}^{n}$, noising operator $\mathcal{N}$, coarse time grid $\{ t_{k}^{c} \}$ and fine time grid $\{ t_{i}^{f} \}$.
        \For{$t_{i} \in \{ t_{i}^{f} \}$}
            \For{$j$ in $range(n)$}
                \State $x_{j}^{f}(t_{i+1}) \gets \mathcal{M}(x_{j}^{f}(t_{i}))$ \Comment{Advance forecast ensemble}
                \State $x_{j}^{a}(t_{i+1}) \gets x_{j}^{f}(t_{i+1})$
            \EndFor
            \If{$t_{i} \in \{ t_{k}^{c}\}$}
                \For{$j$ in $range(n)$}
                    \State $y_{j}(t_{i}) \gets \mathcal{N}(\mathcal{T}(t_{i}))$ \Comment{Generate observation ensemble}
                    \State $x_{j}^{a}(t_{i}) \gets \mathcal{F}(x_{j}^{a}(t_{i}), y_{j}(t_{i}))$ \Comment{Perform KF update}
                \EndFor
            \EndIf
        \EndFor
    \end{algorithmic}
\end{algorithm}

A key point to recognize in this framework is that the initial condition of the ensemble \textit{need not} be the initial condition of the trajectory. This proves particularly useful when one has access to observations of a trajectory but does not know the initial condition corresponding to that trajectory \cite[Chapter 8]{Ahmed20}.

It is important to point out that the ensemble Kalman filter is a Monte-Carlo based extension of the classic Kalman filter \cite{Evensen03}. Under the assumption that all dynamics and observations are linear and all distributions are Gaussian, then the Kalman filter exactly solves the Bayesian filtering problem \citep{Grooms22,Katzfuss16}. One such adjustment possible, as in \cite[19]{Evensen03}, is to replace $R$ with the sample covariance of the errors.

To understand the necessity of the Gaussianity assumption, let us follow a similar analysis carried out in \cite{Grooms22}. Suppose the analysis and forecast ensembles at time $t_{i}$ are sampled from the random variables $X^{a}_{t_{i}}$ and $X^{f}_{t_{i}}$, respectively, and that the observation $y$ is sampled from $Y_{t_{i}}$. $X^{a}_{t_{i}}$ is often defined by
\begin{equation} \label{Eq: Ideal Analysis Update}
    X^{a}_{t_{i}} := \mathbb{E}[X^{f}_{t_{i}} | Y_{t_{i}}].
\end{equation}
Under the assumption that the joint random variable $(X_{t_{i}}^{f},Y_{t_{i}})$ is Gaussian, then \eqref{Eq: Ideal Analysis Update} reduces to
\begin{equation}
    X^{a}_{t_{i}} = \mathbb{E}[X^{f}_{t_{i}}] + Cov(X_{t_{i}}^{f}, Y_{t_{i}}) Cov(Y_{t_{i}})^{-1}(Y_{t_{i}} - \mathbb{E}[Y_{t_{i}}]).
\end{equation}

Let $C_{y}$ be the sample covariance matrix of the observation ensemble $\{ y_{j} \}$ and let $C_{xy}$ be the sample covariance matrix of the joint ensemble $\{ x_{j}^{f}, y_{j} \}$. \citet[10]{Grooms22} shows that the Kalman filter update \ref{Eq: Kalman update} can be expressed as
\begin{equation} \label{Eq: Conditional Gaussian update}
    x_{j}^{a} = x_{j}^{f} + C_{xy}C_{y}^{-1}(y-y_{j}),
\end{equation}
where $y$ is a reference observation vector. Under the assumptions for EnKF (i.e., \eqref{Assumption: Linear observations}), \cite{Grooms22} shows that \eqref{Eq: Conditional Gaussian update} immediately recovers the classic EnKF update formula \eqref{Eq: Kalman update}. Observe that \eqref{Eq: Conditional Gaussian update} does not explicitly require a well-defined linear operator $H$, in contrast to \eqref{Eq: Kalman update}. As such, one may still apply \eqref{Eq: Conditional Gaussian update} when the perturbation is nonlinear. Nevertheless, \eqref{Eq: Conditional Gaussian update} only achieves the ideal update \eqref{Eq: Ideal Analysis Update} when $(X,Y)$ is jointly Gaussian, an assumption that clearly does not always hold.

A technique to sidestep the Gaussian assumption, introduced in \cite{Zhou11}, is to explicitly transform all variables to a Gaussian distribution. In particular, before performing the Kalman filter update \eqref{Eq: Kalman update}, both observations and the ensemble are mapped to a sample that is normally distributed. Such a model is called the normal-score ensemble Kalman filter (NS-EnKF). The normal score transform is a kind of \textit{Gausian anamorphosis} technique, wherein a random scalar variable $X$ whose law is non-atomic can be transformed to a random Gaussian variable $\Hat{X}$ by the mapping
\begin{equation}\label{Def: NS-Transform}
    \Hat{X} := \Phi^{-1} (F_{X}(X)),
\end{equation}
where $F_{X}$ is the cumulative distribution function (cdf) of $X$ and $\Phi$ is the cumulative distribution function for the standard normal distribution. 

NS-EnKF as proposed in \cite{Zhou11} only normal-score transforms the ensemble and adds Gaussian noise to the observation. Indeed, transforming the observations introduces an inconsistency in the current EnKF framework - the matrices $R$ and $H$ no longer become well defined. If the observation is transformed after its perturbation (as in our case), then what does the covariance matrix $R$ become in the latent space? We know that the observations should be transformed to a Gaussian distribution in order to satisfy the assumptions of the Kalman filter, but the update \eqref{Eq: Kalman update} does not make sense. The authors of \cite{Hansen24} normal-score transform the observations as well, but they add Gaussian noise in the latent space. As such, they cannot account for non-Gaussian observation errors. While this inconsistency is not discussed in either \cite{Zhou11} or \cite{Hansen24}, we do believe it is worth acknowledging. In many practical applications, the user does not know the perturbation operator $\mathcal{N}$. It is important to devise an algorithm that treats $\mathcal{N}$ as a \textit{black box} operator and does not rely on it in any latent space.

\citet{Grooms22} sidesteps this issue by normal-score transforming both the forecast $\{ x_{j}^{f} \}$ and observations ensemble $\{ y_{j} \}$ (that is, apply \eqref{Def: NS-Transform} on each dimension for each element in either ensemble) and then applying \eqref{Eq: Conditional Gaussian update} in the \textit{latent space}. Such models are referred to as `GA-PL' and `GA-KDE' in that paper, but we will simply use a KDE-based technique and refer to it as NS-EnKF.

We shall denote the normal-score transform on the ensemble by $\Psi_{x}$. The observation cdf is created using both the perturbed observations $\{ y_{j} \}$ and a \textit{perturbed} ensemble $\{ \mathcal{N}(x_{j}^{f}) \}$. We shall denote the normal-score transform on the observations by $\Psi_{y}$. The intuition behind this approach is that ensemble states transformed by the observation operator $\mathcal{N}$ should come from the same distribution as the observations $\{ y_{j} \}$. This agrees with the method described by \cite{Grooms22} and lets us make a more robust normal-score transform. Nevertheless, the normal-score transform can only be applied on a per-dimension basis. By its very definition, it is a univariate transform.

In our implementation, we took $y$ to be the original observation vector at that time. One can further modify \eqref{Eq: Conditional Gaussian update} by incorporating \textit{localization} and \textit{inflation}. For the latter, we multiply the forecast ensemble by a positive inflation factor $r$ before assimilation
\begin{equation}
    x_{j}^{f}(t) \gets \Bar{x}^{f}(t) + r(x_{i}^{f}(t) - \Bar{x}^{f}).
\end{equation}
We define $\Bar{x}^{f}(t)$ to be the forecast mean, that is
\begin{equation}
    \Bar{x}^{f}(t) = \frac{1}{n} \sum_{j}^{n}x_{j}^{f}(t).
\end{equation}
Please refer to \cite{Grooms22} for a more complete discussion of localization. For the former, define a \textit{localization matrix} $C$ and rather than using the covariance matrices $C_{xy}$ and $C_{y}$ directly, we use the update formula
\begin{equation}\label{Eq: Actual CG Update}
    x_{j}^{a} = x_{j}^{f} + (L \odot C_{xy})(L \odot C_{y})^{-1}(y-y_{j}),
\end{equation}
where $\odot$ refers to the Hadamard (entry-wise) product. Choosing a localization matrix is a nontrivial task, but is outside the scope of this paper. Please refer to \cite{Janjic11} for a complete review of localization. For simplicity, let us denote the operation on the right hand side of \eqref{Eq: Actual CG Update} with $\mathcal{F}_{c}$, for `conditional Gaussian update'.

\begin{algorithm}
    \caption{NS-EnKF}  \label{Alg: NS-EnKF}
    \begin{algorithmic} 
        \State Given reference trajectory $\mathcal{T}$, dynamical model $\mathcal{M}$, initial ensemble $\{ x_{j}(0) \}_{j=1}^{n}$, noising operator $\mathcal{N}$, coarse time grid $\{ t_{k}^{c} \}$, inflation factor $r$, and fine time grid $\{ t_{i}^{f} \}$.
        \For{$t_{i} \in \{ t_{i}^{f} \}$}
            \For{$j$ in $range(n)$}
                \State $x_{j}^{f}(t_{i+1}) \gets \mathcal{M}(x_{j}^{f}(t_{i}))$ \Comment{Advance forecast ensemble}
                \State $x_{j}^{a}(t_{i+1}) \gets x_{j}^{f}(t_{i+1})$
            \EndFor
            \If{$t_{i} \in \{ t_{k}^{c}\}$}
                \For{$j$ in $range(n)$} \Comment{Generate observation ensemble}
                    \State $y_{j}(t_{i}) \gets \mathcal{N}(\mathcal{T}(t_{i}))$
                \EndFor
                \State $\{ \hat{x}_{j}^{f}(t_{i}) \} \gets \Psi_{x}(\{ x_{j}^{f}(t_{i}) \})$ \Comment{Normal-score transform the ensemble}
                \State $\{ \hat{y}_{j}(t_{i}) \} \gets \Psi_{y}(\{ x_{j}^{f}(t_{i}) \}, \{ y_{j}(t_{i}) \})$ \Comment{Normal-score transform the observations}
                \For{$j$ in $range(n)$} 
                    \State $x_{j}^{f}(t) \gets \Bar{x}^{f}(t) + r(x_{i}^{f}(t) - \Bar{x}^{f})$ \Comment{Localize} 
                    \State $\hat{x}_{j}^{a}(t_{i}) \gets \mathcal{F}_{c}(\hat{x}_{j}^{f}(t_{i}), \hat{y}_{j}(t_{i}))$ \Comment{Perform conditional Gaussian update}
                \EndFor
                \State $\{ x_{j}^{a}(t_{i})\} \gets \Psi_{x}^{-1}(\{ \hat{x}_{j}^{a}(t_{i})\})$ \Comment{Inverse normal-score transform analysis ensemble}
            \EndIf
        \EndFor
    \end{algorithmic}
\end{algorithm}

As discussed in \cite{Grooms22}, this update can also be used with the transformed ensemble and observations, as seen in  Algorithm~\ref{Alg: NS-EnKF}. After performing the update, one can invert the normal score transform and recover an analysis ensemble in the state space. Crucially, localization and inflation are performed on the latent space.

It should also be noted that the update in \eqref{Eq: Conditional Gaussian update} no longer requires the observation matrix $H$. According to \cite{Grooms22}, it is applicable for a wide range of observation operators $\mathcal{N}$, instead of a strictly linear one. As seen in \cite{Grooms22}, both EnKF and NS-EnKF can be generalized for nonlinear $\mathcal{N}$ using the conditional Gaussian update \eqref{Eq: Conditional Gaussian update}. Algorithm~\ref{Alg: NS-EnKF} details NS-EnKF using the conditional Gaussian update, but the CG-EnKF is implemented analogously, except without normal-score transformations.

\section{Score Filter}
As previously discussed, there are many limitations in the vanilla EnKF approach detailed in Algorithm~\ref{Alg: EnKF}. \cite{Bao24} uses these limitations to justify Score Filter (SF). SF is a kind of particle filter, which in contrast to Kalman filters, aims to solve the filtering problem by sequential importance sampling \citep{Grooms22}. In particular, classic particle filters construct an empirical measure that weakly converges to the correct posterior when the ensemble size tends to infinity \cite{Grooms22}. However, as discussed in both \cite{Grooms22} and \cite{Bao24}, the classical particle filter attains prohibitively slow convergence in high dimensional data, thereby suffering from the `curse of dimensionality'.

SF aims to avoid the curse of dimensionality by using a diffusion model to sample from the Bayesian posterior \cite{Bao24}. Indeed, diffusion models have proven remarkably successful in sampling from various high dimensional data. We encourage the interested reader to consult \cite{Yang23} for a survey on diffusion models and their success in solving a vast array of generative modelling problems. One particular class of diffusion models are score-based, in that the model learns a score function with which to evolve particles through a stochastic differential equation (SDE) \citep{Song21}. SF uses this kind of diffusion model for their posterior sampling \citep{Bao24}. We use the rest of this section to summarize the workflow of SF and discuss its strengths and limitations. We also note a notational difference from \cite{Bao24}. Keep in mind that analysis and forecast in this paper refers to the posterior and prior in \cite{Bao24}.

For simplicity, let us assume $\{t_{i}^{c}\} = \{t_{i}^{f}\} = \{t_{i}\}$, that is, we perform a filtering step at each time point in the discretization. As in the Kalman filtering case, let us start with a forecast ensemble $\{ x_{j}^{f}(t_{i}) \} \sim X^{f}_{t_{i}}$ produced by the model $\mathcal{M}$ from an analysis ensemble $\{ x_{j}^{a}(t_{i-1})\} \sim X^{a}_{t_{i-1}}$ and observation $y(y_{i}) \sim Y_{t_{i}}$. We would like to obtain $X^{a}_{t_{i}}$ using the ideal update \eqref{Eq: Ideal Analysis Update}. However, instead of directly computing $\mathbb{E}[X^{f}_{t_{i}}|Y_{t_{i}}]$, particle filters to obtain the density of $X_{t_{i}}^{a}$, dubbed the \textit{posterior density}. The \textit{prior} density is that corresponding to $X_{t_{i}}^{f}$. In classic particle filters, the prior density is often merely the empirical density of the forecast ensemble. In SF, the forecast (prior) and analysis (posterior) densities are implicitly obtained using a score-based diffusion model. The corresponding score functions are denoted by $S_{t_{i}}^{f}$ and $S_{t_{i}}^{a}$, respectively.

Given a forecast ensemble $\{ x_{j}^{f}(t_{j+1})\}$, one can train a diffusion model to learn a score function $S_{t_{i}}^{f}$. In particular, using this score function, it is possible to generate an \textit{arbitrary} number of realizations of $X_{t_{i}}^{f}$, assuming ideal training. For our purpose, we will denote $\mathcal{L}_{t_{i}}^{F}$ as the \textit{forward} SDE operator for the score-based diffusion model (please consult \cite{Bao24} for the specific form of the forward-backward SDEs used). Let $\{Z_{\tau}^{t_{i}}\}$ be forward evolution process, that is, it satisfies
\begin{equation} \label{Eq: Forward Diffusion Model SDE}
    \begin{aligned}
        dZ_{\tau}^{t_{i}} &= \mathcal{L}^{F}_{t_{i}}(Z_{\tau}^{t_{i}}, S_{t_{i}}^{f}(Z_{\tau}^{t_{i}}, \tau; \theta)) \\
        Z_{0}^{t_{i}} &= X^{f}_{t_{i}}, \\
    \end{aligned}    
\end{equation}
where $Z_{1}^{t_{i}} \equiv Z_{1}$ is some pre-specified random variable with a known distribution. In the case of SF, $Z_{1}$ is taken to be a standard normal random variable. While it is known that diffusion models do not always attain $Z_{1}$ through the forward diffusion \eqref{Eq: Forward Diffusion Model SDE} \citep{Chen23}, SF assumes that it does (which is a common assumption). Recall that under ideal training, the score function should satisfy
\begin{equation}
    S_{t_{i}}^{f}(z, \tau; \theta) \approx \nabla_{z} \log \rho_{\tau}^{t_{i}}(z),
\end{equation}
where $\rho_{\tau}^{t_{i}}$ is the density of $Z_{\tau}^{t_{i}}$.

\begin{algorithm}
    \caption{SF} \label{Alg: SF}
    \begin{algorithmic}
        \State Given reference trajectory $\mathcal{T}$, dynamical model $\mathcal{M}$, initial ensemble $\{ x_{j}(0) \}_{j=1}^{n}$, noising operator $\mathcal{N}$ and time grid $\{ t_{i} \}$
        \State Train $S_{0}^{a}$ given the initial ensemble $\{ x_{j}^{a}(t_{0})\}$.
        \For{$t_{i} \in \{t_{i}\}$}
            \State $\{ x_{j}^{a}(t_{0}) \} \gets S_{t_{i}}^{a}$. \Comment{Use current score function to generate analysis ensemble.}
            \For{$j$ in $range(n)$}
                \State $x_{j}^{f}(t_{i+1}) \gets \mathcal{M}(x_{j}^{a}(t_{i}))$ \Comment{Advance analysis ensemble with $\mathcal{M}$.}
            \EndFor
            \State $S_{t_{i}}^{f} \gets \{ x_{j}^{f}(t_{i})\}$ \Comment{Use forecast ensemble to train prior score function.}
            \State $y(t_{i}) \gets \mathcal{N}(\mathcal{T}(t_{i}))$
            \State $S_{t_{i}}^{a} \gets (S_{t_{i}}^{f}, y(t_{i}))$ \Comment{Use prior score function with observation to obtain posterior score function.}
        \EndFor
    \end{algorithmic}
\end{algorithm}

Samples of $X_{t_{i}}^{f}$ are produced by following the corresponding \textit{backward} SDE.
SF then uses a clever update technique \cite[Equation 20]{Bao24} on the score function $S_{t_{i}}^{f}$ to obtain $S_{t_{i}}^{a}$. To be precise, the new score function is defined by
\begin{equation} \label{Eq: SF Update}
    S^{a}_{t_{i}}(Z_{\tau}, \tau; \theta) := S^{f}_{t_{i}}(Z_{\tau}, \tau; \theta) + h(\tau) \nabla_{z} L_{t_{i}}(Z_{\tau}, \tau; y(t_{i})),
\end{equation}
where $L(Z_{\tau}; y(t_{i}))$ is a \textit{score} likelihood function (that is, proportional to a $\log$ likelihood function). With $S^{a}_{t_{i}}$, one can now generate an arbitrary collection of realizations of $X_{t_{i}}^{a}$. However, it should be pointed out that SF was derived in the case of additive \textbf{Gaussian} perturbations. That is, the perturbation operator takes the form
\begin{equation}
    \mathcal{N}(x) = g(x) + \epsilon
\end{equation}
for some $\epsilon \sim N(0,R)$ and smooth function $g$. For this case, \cite{Bao24} use the score likelihood
\begin{equation} \label{Eq: SF Likelihood}
    L^{(i)}_{t_{i}}(z^{(i)}, \tau; y^{(i)}) = \frac{g(z^{(i)})-y^{(i)}(t_{i})}{\sigma^{2}} \frac{d}{dz^{(i)}} g(z^{(i)}) ReLU(1-2\tau), \quad t_{i} \neq 0,
\end{equation}
in their simulations, where superscripts denote the $i$th element of each vector and $z \sim Z_{\tau}$. When $t_{i}=0$, then $L_{t_{i}}(z^{(i)}, \tau; y^{(i)})$ = 0. It should be noted that \cite{Bao24} did not give an explicit formula for this log likelihood in their paper. Rather, this function was gleamed through a careful consideration of the code used to generate their figures\footnote{Please refer to \url{https://github.com/zezhongzhang/Score-based-Filter} for the code implementation of \cite{Bao24}.}. We believe that a proper justification for such a likelihood function is lacking and this lack of clarity is an apparent drawback of SF.

A pseudo-code implementation is given in Algorithm~\ref{Alg: SF}. Unfortunately, this expanded generality comes at the cost of both training a neural network and solving an SDE at each time step. As we shall see in the next section, this results in a far more expensive algorithm as compared to CG-EnKF or NS-EnKF.

\section{Experimental Results}
\textbf{SF versus EnKF.} We test the previously discussed methodologies on assimilating data with the $40$-dimensional Lorenz-96 system. In other words, the true trajectory, $\mathcal{T}(t)$ is the solution of the system of equations
\begin{equation} \label{Eq: L-96}
    \begin{aligned}
        \frac{d x_{j}}{dt} &= (x_{j+1}(t) - x_{j-2}(t)) x_{i-1}(t) + 8, \quad i = 1,2,\cdots 40, \\
        x_{-1}(t) &= x_{39}(t), \\
        x_{0}(t) &= x_{40}(t), \\
        x_{41}(t) &= x_{1}(t).
    \end{aligned}
\end{equation}
Similar to \cite{Bao24}, we experiment with a cubic observation model, that is, the perturbation operator $\mathcal{N}(x)$ is
\begin{equation} \label{Eq: Cubic Perturbations}
    \mathcal{N}_{cub}(x) := x^{3} + \epsilon, \quad \epsilon \sim N(0,1),
\end{equation}
where the exponentiation occurs element-wise. We also take a linear observation model for further comparison
\begin{equation}
    \mathcal{N}_{lin}(x) = x + \epsilon, \quad \epsilon \sim N(0,1).
\end{equation}

We assimilate perturbed trajectories from \eqref{Eq: L-96} using CG-EnKF, NS-EnKF, and SF. In every experiment, we generate a trajectory from \eqref{Eq: L-96} for $1$ second of simulation time. We discretize the time interval into $100$ equally spaced time steps and filter at each time step, for simplicity, let us refer to the discretization by $\{ t_{i} \}$. The goal is to directly compare the assimilation algorithms using classical statistics (that is, CG-EnKF and NS-EnKF) against SF. It is trivially true that vanilla EnKF cannot assimilate trajectories with perturbations of the form \eqref{Eq: Cubic Perturbations}, as per our previous discussion. While \cite{Bao24} uses that fact to propose SF, we wish to examine whether CG-EnKF and NS-EnKF can solve this problem.

In both Kalman filter algorithms, we use the localization matrix
\begin{equation}
    C_{ij} := \exp \left\{ -\frac{1}{2} \left( \frac{d_{ij}}{d} \right)^{2} \right\},
\end{equation}
where 
\begin{equation}
    d_{ij} = \min \{|i-j|, 40-|i-j| \}
\end{equation}
and $d$ is the localization radius, following the same convention as \cite{Grooms22}. We took $d = 1$ in our experiments. We also took inflation factor $r = 1.05$ for both filtering implementations. The goal of this experiment is not to obtain the best possible CG-EnKF or NS-EnKF implementation to solve the problem. Rather, we want to use a n\"{a}ive implementation of CG-EnKF and NS-EnKF and compare their performance against that of SF on a well studied benchmark problem. We base our implementations of CG-EnKF and NS-EnKF off the publicly available implementations used in \cite{Grooms22}.

For SF, we train the diffusion model for $1000$ epochs at each assimilation cycle. That is, the prior score function $s_{t_{i+1}|t_{i}}$ is trained for $1000$ epochs for each $t_{i}$ in the discretization $\{ t_{i} \}$. We use the exact same score diffusion model as in \cite{Bao24}, the reason being we wish to provide a fair comparison between SF and the Kalman filters in consideration. 

The first statistic we use to validate results is the root mean squared error (RMSE). At some time $t \in \{ t_{i} \}$, we compute forecast RMSE (FRMSE) by the formula
\begin{equation}
    \sqrt{\frac{1}{40} \sum_{j=1}^{40} (x_{j}^{f}(t) - \mathcal{T}(t))^{2}} .
\end{equation}
For some time $t \in \{ t_{i}^{c} \}$, we compute analysis RMSE (ARMSE) by
\begin{equation}
    \sqrt{\frac{1}{40} \sum_{j=1}^{40} (x_{j}^{a}(t) - \mathcal{T}(t))^{2}}.
\end{equation}

\begin{figure}[h]
    \centering
    \begin{subfigure}{0.32\textwidth}
        \includegraphics[width=\linewidth, height=6cm, keepaspectratio]{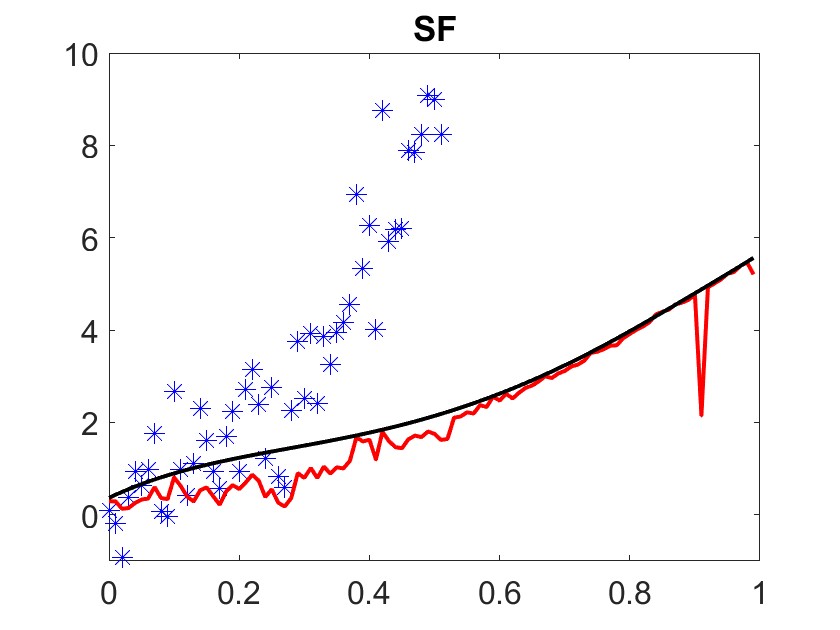} 
        \label{fig:subim1}
    \end{subfigure}
    \begin{subfigure}{0.32\textwidth}
        \includegraphics[width=\linewidth, height=6cm, keepaspectratio]{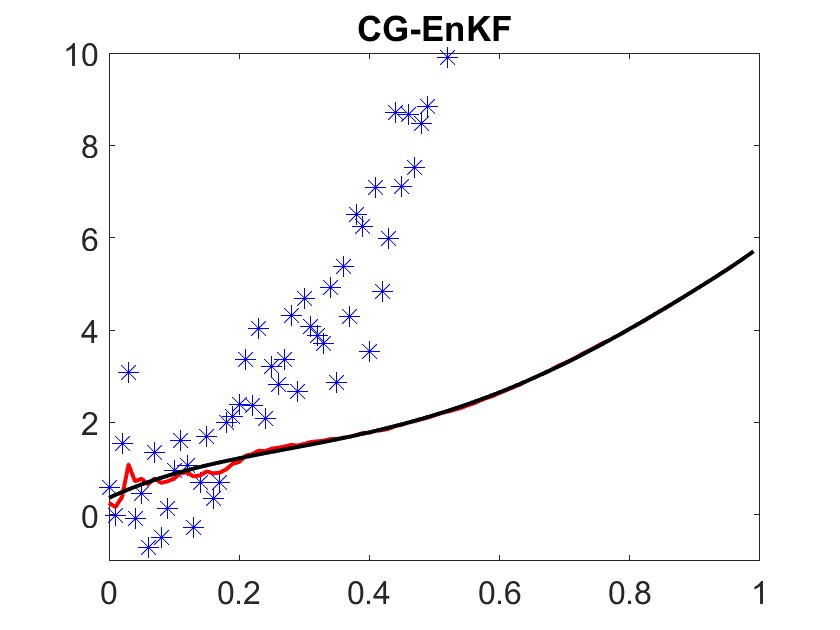}
        \label{fig:subim2}
    \end{subfigure}
    \begin{subfigure}{0.32\textwidth}
        \includegraphics[width=\linewidth, height=6cm, keepaspectratio]{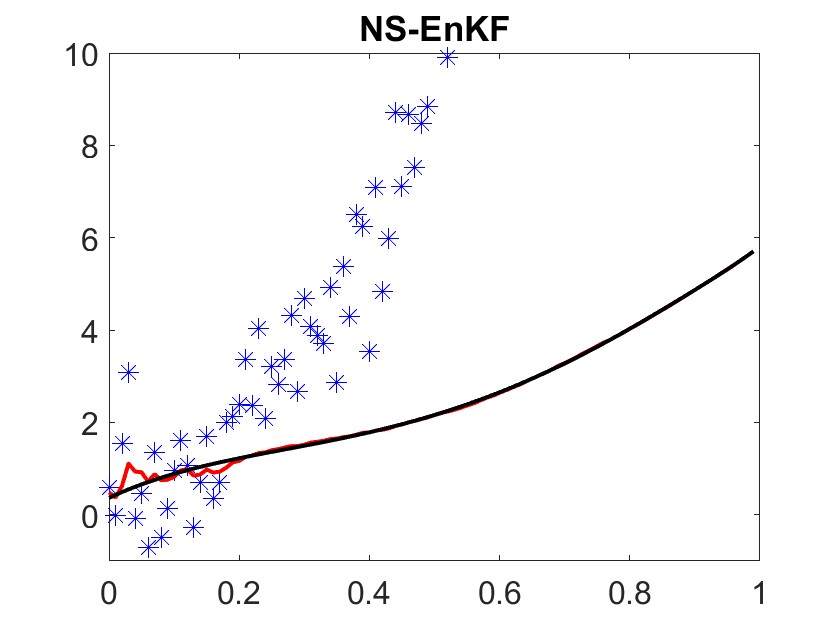}
        \label{fig:subim3}
    \end{subfigure}
    
    \begin{subfigure}{0.32\textwidth}
        \includegraphics[width=\linewidth, height=6cm, keepaspectratio]{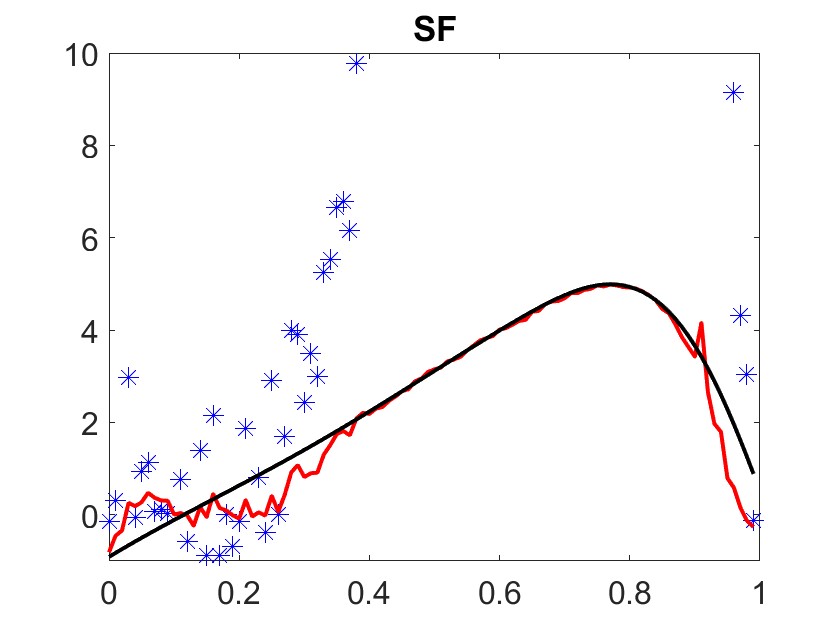} 
        \label{fig:subim4}
    \end{subfigure}
    \begin{subfigure}{0.32\textwidth}
        \includegraphics[width=\linewidth, height=6cm, keepaspectratio]{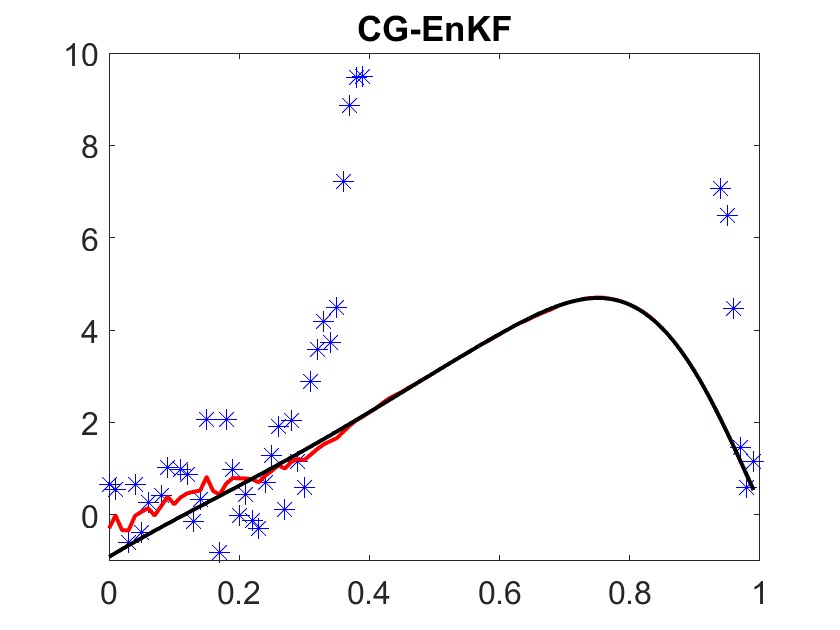}
        \label{fig:subim5}
    \end{subfigure}
    \begin{subfigure}{0.32\textwidth}
        \includegraphics[width=\linewidth, height=6cm, keepaspectratio]{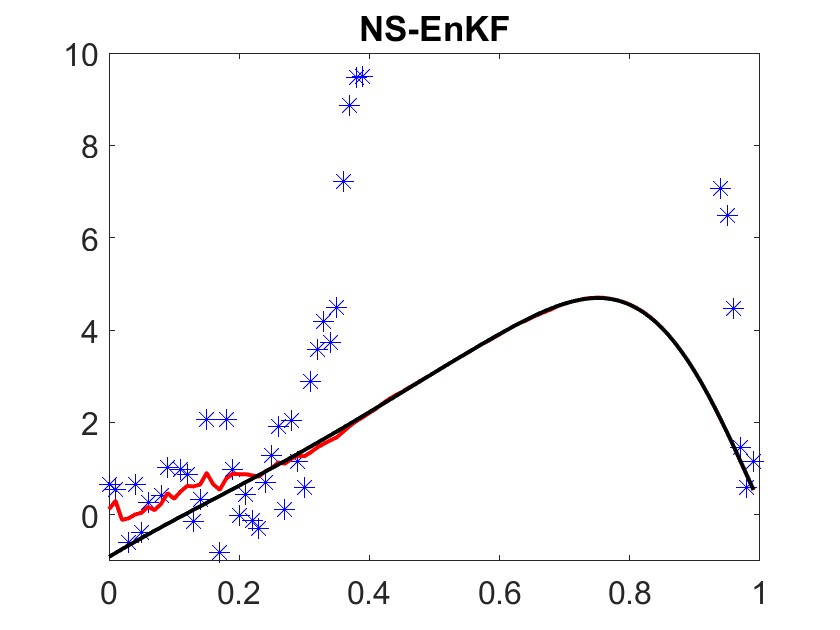}
        \label{fig:subim6}
    \end{subfigure}
    
    \begin{subfigure}{0.32\textwidth}
        \includegraphics[width=\linewidth, height=6cm, keepaspectratio]{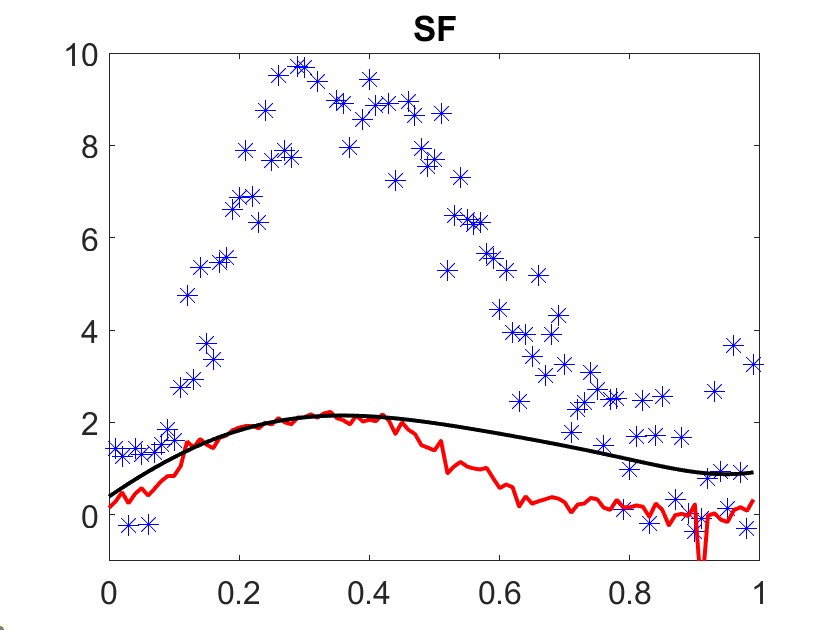} 
        \label{fig:subim7}
    \end{subfigure}
    \begin{subfigure}{0.32\textwidth}
        \includegraphics[width=\linewidth, height=6cm, keepaspectratio]{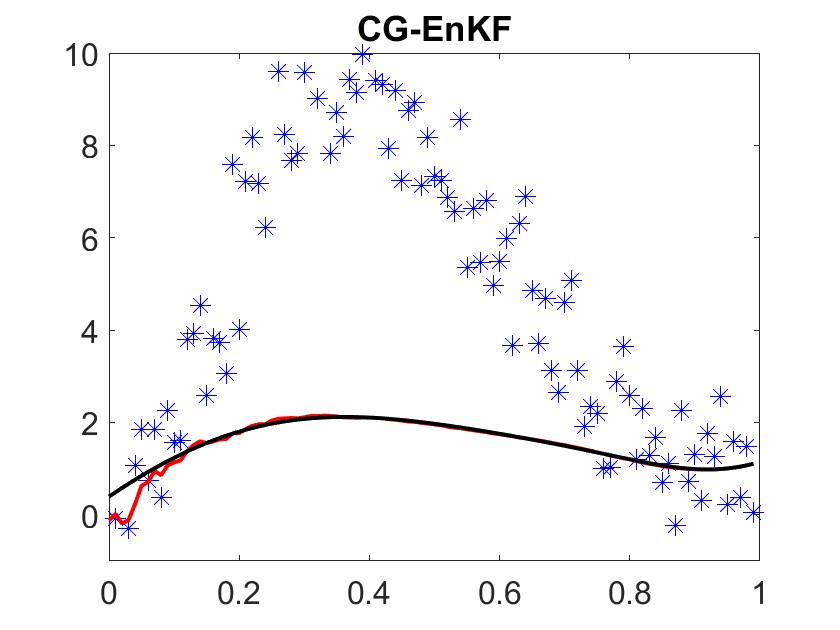}
        \label{fig:subim8}
    \end{subfigure}
    \begin{subfigure}{0.32\textwidth}
        \includegraphics[width=\linewidth, height=6cm, keepaspectratio]{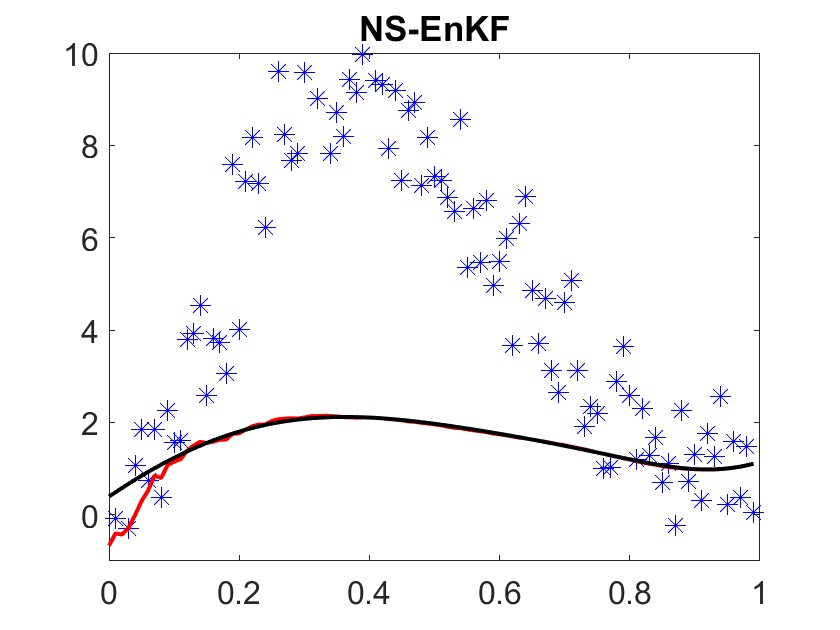}
        \label{fig:subim9}
    \end{subfigure}    
    \caption{Assimilated trajectories (red) against true trajectory (black) and cubic perturbed observations (blue stars) for SF (left), CG-EnKF (middle), and NS-EnKF (right), for L96 along dimensions $1$ (top), $20$ (middle), and $40$ (bottom).}
    \label{fig:Assimilated images}
\end{figure}

\begin{table}[h]
    \centering
    \begin{tabular}{ |p{1.8cm}|p{1.5cm}|p{1.5cm}|p{1.5cm}|p{1.5cm}|p{1.5cm}|}
        \hline
        Experiment & Mean FCRPS & Mean ACRPS & FRMSE & ARMSE & Time (s) \\
        \hline
        CG-EnKF & 0.0365 & 0.0343 & 0.0706 & 0.0702 & 4.6840 \\ 
        NS-EnKF & 0.0444 & 0.0421 & 0.0868 & 0.0865 & 33.982\\ 
        SF & 0.1467 & 0.1656 & 0.0631 & 0.0631 & 1119.6\\ 
        \hline 
    \end{tabular}
    \caption{Assimilation Results on $1$s of L-96 with Cubic Observations}
    \label{Table:Cubic SF vs EnKF Results}
\end{table}

\begin{table}[h]
    \centering
    \begin{tabular}{ |p{1.8cm}|p{1.5cm}|p{1.5cm}|p{1.5cm}|p{1.5cm}|p{1.5cm}|}
        \hline
        Experiment & Mean FCRPS & Mean ACRPS & FRMSE & ARMSE & Time (s) \\
        \hline
        CG-EnKF & 0.1384 & 0.1370 & 0.2369 & 0.2376 & 4.1376 \\ 
        NS-EnKF & 0.1364 & 0.1349 & 0.2330 & 0.2335 & 42.125\\ 
        SF & 0.1766 & 0.2096 & 0.0471 & 0.0471 & 1157.3\\ 
        \hline 
    \end{tabular}
    \caption{Assimilation Results on $1$s of L-96 with Linear Observations}
    \label{Table:Linear SF vs EnKF Results}
\end{table}

The second statistic we consider is the continuous ranked probability score (CRPS), for both the forecast states and the analysis states. At a high level, the CRPS measures the quality of a probabilistic estimate. A CRPS of $0$ for a given probabilistic estimate is the best possible score attainable, although not even the true Bayesian posterior will have a CRPS of $0$ \cite{Grooms22}. The interested reader is encouraged to consult \cite{Hersbach20} for a thorough explanation of the CRPS. 

Lastly, we measure the wallclock time each implementation took to run on a standard laptop CPU. It should be noted that \cite{Bao24} did not include compute time in their results and there may be discrepancies based on code optimization and hardware specifications. 


\begin{figure}[h!]
    \centering
    \begin{subfigure}{0.32\textwidth}
        \includegraphics[width=\linewidth, height=6cm, keepaspectratio]{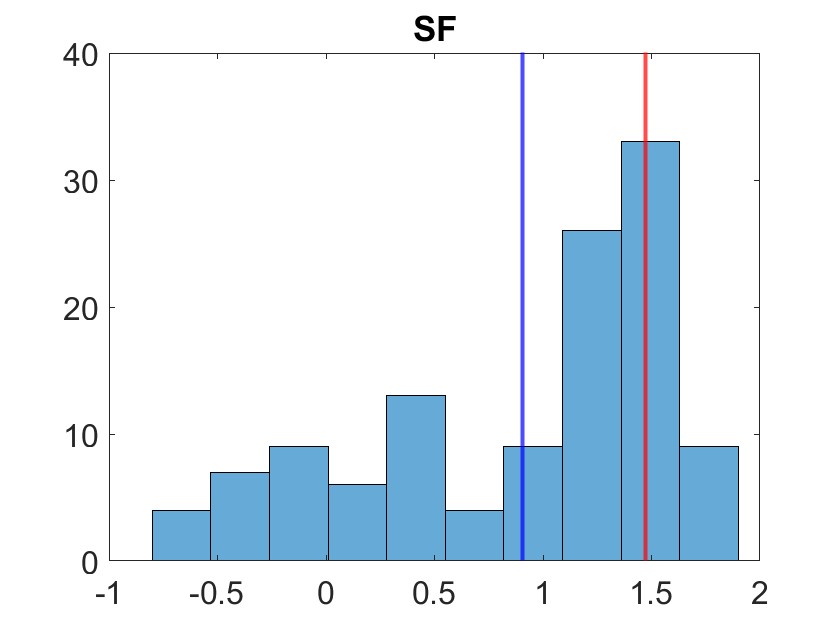} 
    \end{subfigure}
    \begin{subfigure}{0.32\textwidth}
        \includegraphics[width=\linewidth, height=6cm, keepaspectratio]{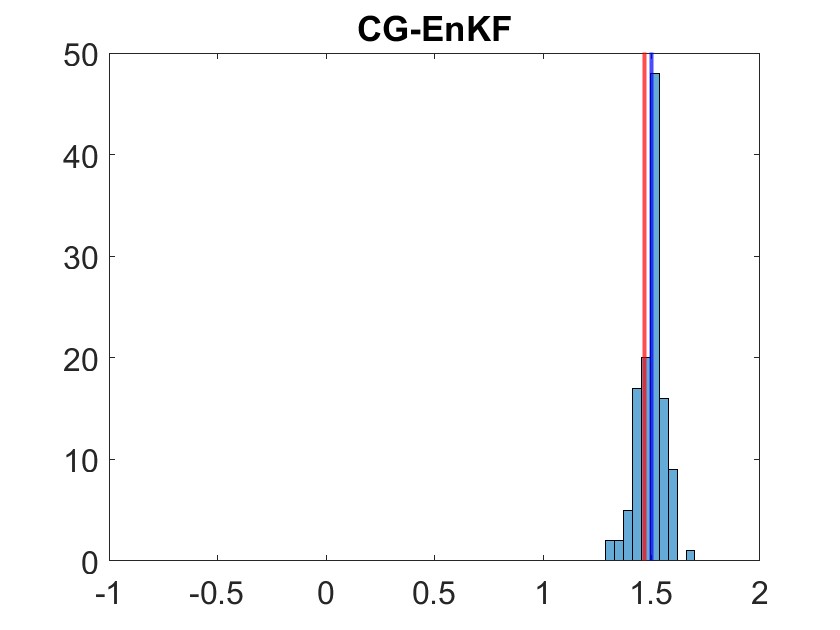}
    \end{subfigure}
    \begin{subfigure}{0.32\textwidth}
        \includegraphics[width=\linewidth, height=6cm, keepaspectratio]{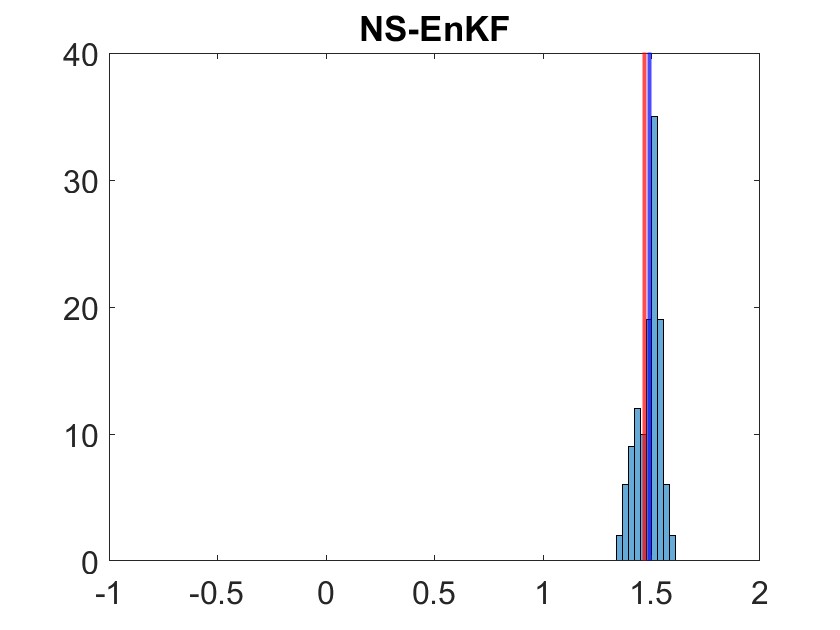}
    \end{subfigure}
    \begin{subfigure}{0.32\textwidth}
        \includegraphics[width=\linewidth, height=6cm, keepaspectratio]{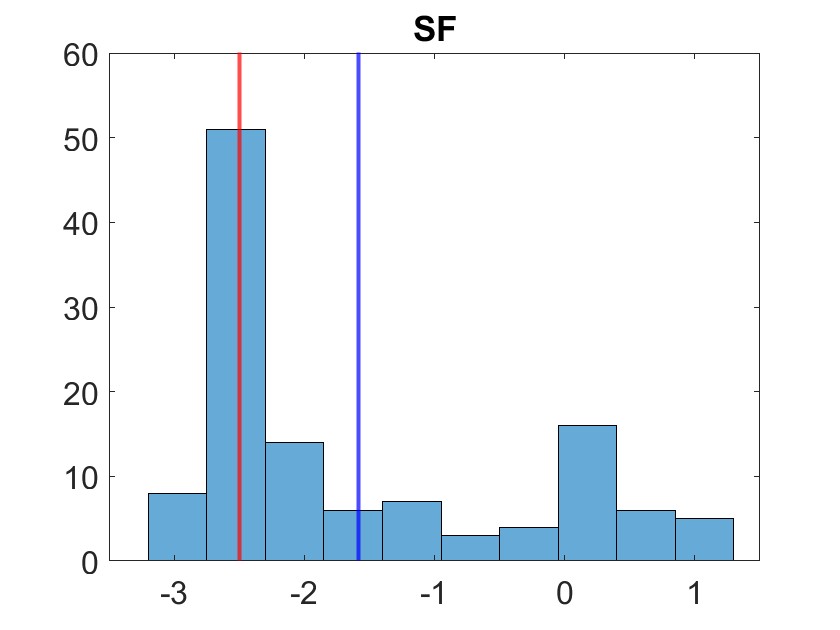} 
    \end{subfigure}
    \begin{subfigure}{0.32\textwidth}
        \includegraphics[width=\linewidth, height=6cm, keepaspectratio]{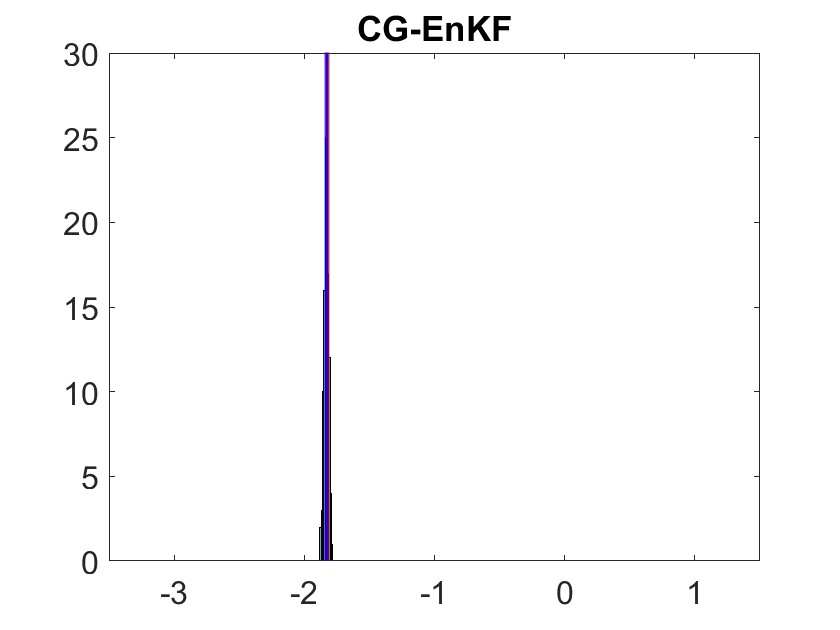}
    \end{subfigure}
    \begin{subfigure}{0.32\textwidth}
        \includegraphics[width=\linewidth, height=6cm, keepaspectratio]{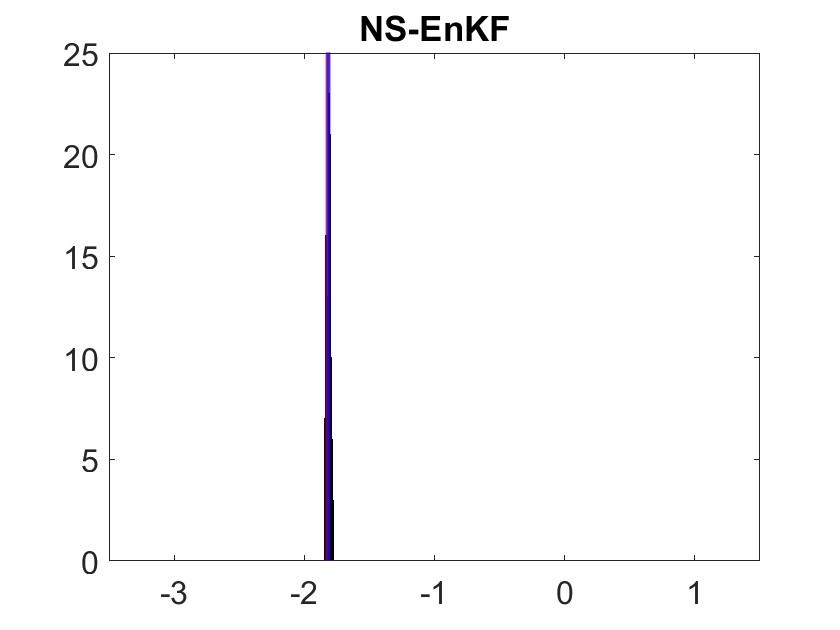}
    \end{subfigure}
    \begin{subfigure}{0.32\textwidth}
        \includegraphics[width=\linewidth, height=6cm, keepaspectratio]{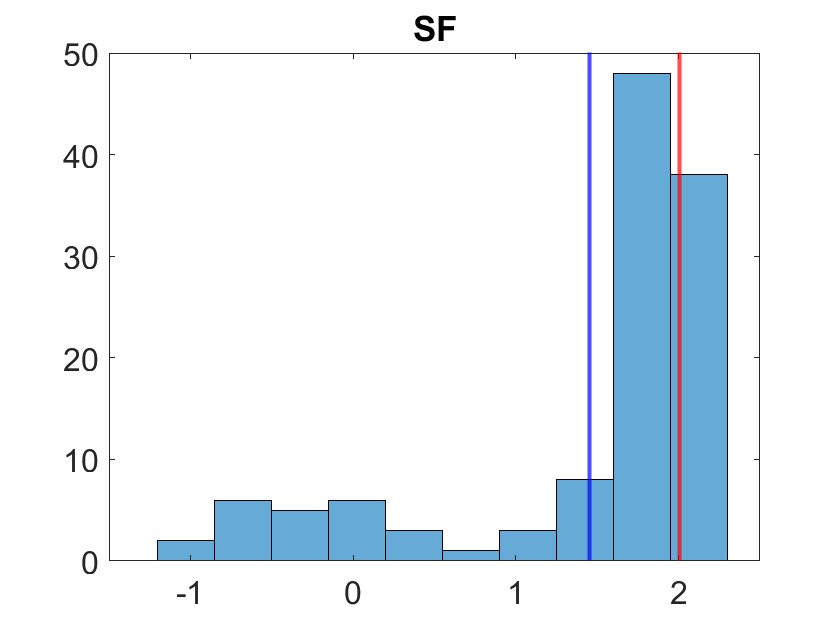} 
    \end{subfigure}
    \begin{subfigure}{0.32\textwidth}
        \includegraphics[width=\linewidth, height=6cm, keepaspectratio]{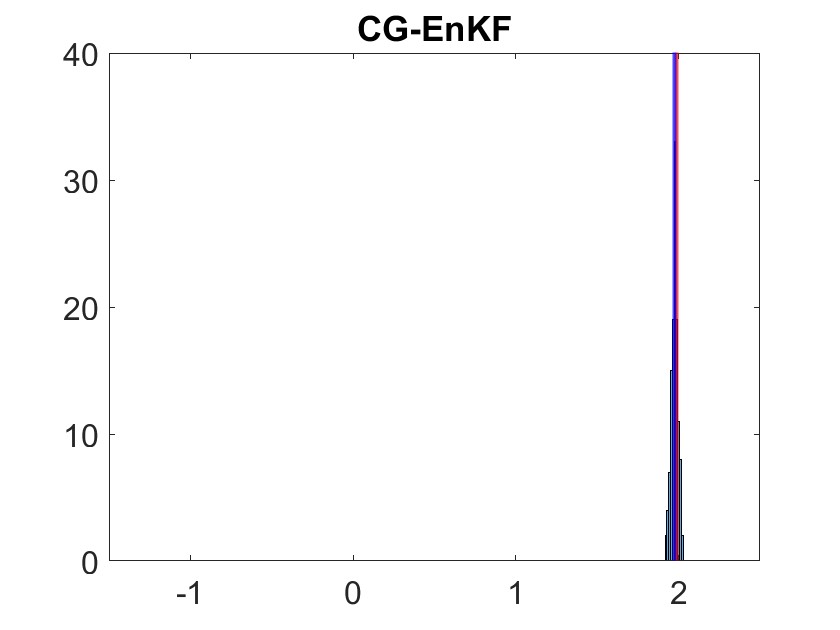}
    \end{subfigure}
    \begin{subfigure}{0.32\textwidth}
        \includegraphics[width=\linewidth, height=6cm, keepaspectratio]{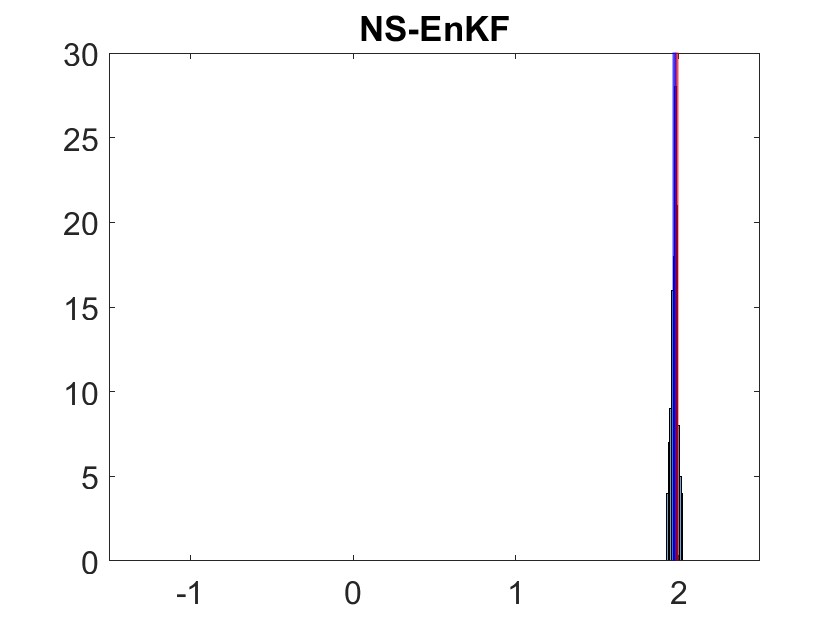}
    \end{subfigure}
    \caption{Histograms of the analysis ensemble of the first dimension at the $30$th (top) time step, the $15$th dimension at the $90$th time step (middle), and the $40$th dimension as the $50$th time step (bottom) for each of the models, with cubic perturbations. The blue line gives the mean of the ensemble and the red line gives the ground truth.}
    \label{Fig: Hists of ensembles against true vals}
\end{figure}

We plot the mean analysis trajectories for the first, $20$th, and $40$th dimensions in \ref{fig:Assimilated images}. The analysis ensemble is also compared against the true value for each model at various times in \ref{Fig: Hists of ensembles against true vals}. From these two figures, we may qualitatively observe stronger performance in CG-EnKF and NS-EnKF over SF. Indeed, the analysis ensemble should resemble a dirac distribution about the true value, which we observe for CG-EnKF and NS-EnKF in \ref{Fig: Hists of ensembles against true vals}. Such a quality is clearly not true for SF in the selected time points.

The quantitative results presented in Table~\ref{Table:Cubic SF vs EnKF Results} validate our qualitative observation. Furthermore, they elucidate the time complexities of the varying algorithms. Indeed, CG-EnKF solved the problem in under three seconds, or roughly $500$ times faster than $SF$. We remark that SF was implemented in Python whereas our CG-EnKF and NS-EnKF implementations are in MATLAB, but MATLAB's slight superiority in matrix operations is not sufficient in itself to account for the vast difference in compute time. It is well known that training neural networks and solving SDEs are expensive, both of which are necessary for \textit{each time step} of SF. Moreoever, we can observe in Tables~\ref{Table:Cubic SF vs EnKF Results} and \ref{Table:Linear SF vs EnKF Results} that the analysis statistics are typically better than the forecast statistics for CG-EnKF and NS-EnKF, whereas for SF, they are often \textit{worse}. 

\textbf{More Experiments with EnKF.} Next, we test the efficacy of our implementations of CG-EnKF and NS-EnKF with a variety of perturbations and noises. With the exception of the cubic observations model, we ran comparative results using the vanilla EnKF as described in Algorithm~\ref{Alg: EnKF}. In Table~\ref{Table: EnKF Results for 5500 assimilation cycles}, this implementation is referred to as `V-EnKF'. We ran experiments on the $40$-dimensional Lorenz-96 system for $5500$ filter steps, with each filter step occurring on every time point along the discretization. We start the ODE trajectory at $0s$ and begin assimilation at $9s$ in ODE time. 

\begin{table}[h!]
\centering
\begin{tabular}{|c|c|c|c|c|c|c|c|}
\hline
Observation & Experiment & FCRPS & ACRPS & FRMSE & ARMSE & Inflation & Time(s) \\ \hline
\multirow{3}{*}{Linear} & CG-EnKF & 0.1904 & 0.1752 & 0.3469 & 0.3163 & 1 & 156.292 \\ \cline{2-8} 
                       & NS-EnKF & 0.1825 & 0.1695 & 0.3325 & 0.2838 & 1.05 & 3012.8 \\ \cline{2-8} 
                       & V-EnKF & 0.1676 & 0.1548 & 0.3094 & 0.2838 & 1 & 181.219 \\ \hline
\multirow{3}{*}{Exponential} & CG-EnKF & 0.1959 & 0.1802 & 0.3556 & 0.3240 & 1 & 215.52 \\ \cline{2-8} 
                       & NS-EnKF & 0.1411 & 0.1307 & 0.2591 & 0.2387 & 1.05 & 2791.6 \\ \cline{2-8} 
                       & V-EnKF & 0.1711 & 0.1575 & 0.3163 & 0.2895 & 1 & 215.52 \\ \hline
\multirow{3}{*}{Bimodal} & CG-EnKF & 0.9627 & 0.9072 & 1.7272 & 1.6185 & 1 & 182.03 \\ \cline{2-8} 
                       & NS-EnKF & 0.7112 & 0.6648 & 1.2878 & 1.1978 & 1 & 2734.7 \\ \cline{2-8} 
                       & V-EnKF & 0.8383 & 0.7880 & 1.5233 & 1.4251 & 1 & 155.08 \\ \hline
\multirow{2}{*}{Cubic} & CG-EnKF & 0.0044 & 0.0040 & 0.0080 & 0.0073 & 1 & 155.08 \\ \cline{2-8} 
                       & NS-EnKF & 0.0036 & 0.0036 & 0.0072 & 0.0066 & 1.05 & 2412.0 \\ \hline
\multirow{1}{*}{Pareto} & NS-EnKF & 0.2556 & 0.2362 & 0.4750 & 0.4381 & 1.05 & 3436.1 \\ \hline 
\end{tabular}
\caption{Results $5500$ assimilation cycles on L-96}
\label{Table: EnKF Results for 5500 assimilation cycles}
\end{table}

\begin{figure}[h]
    \centering
    \includegraphics[width=0.5\linewidth]{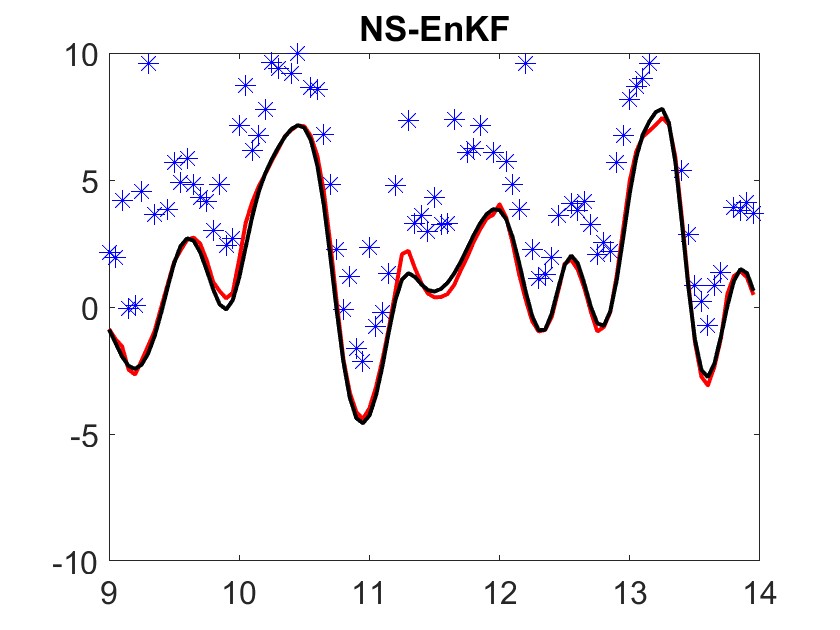}
    \caption{NS-EnKF assimilation of the first $100$ assimilation cycles of the first dimension of L-96 with Pareto additive noise.}
    \label{fig: NSEnKF Pareto}
\end{figure}

\begin{figure}[h!]
    \centering
    \begin{subfigure}{0.32\textwidth}
        \includegraphics[width=\linewidth, height=6cm, keepaspectratio]{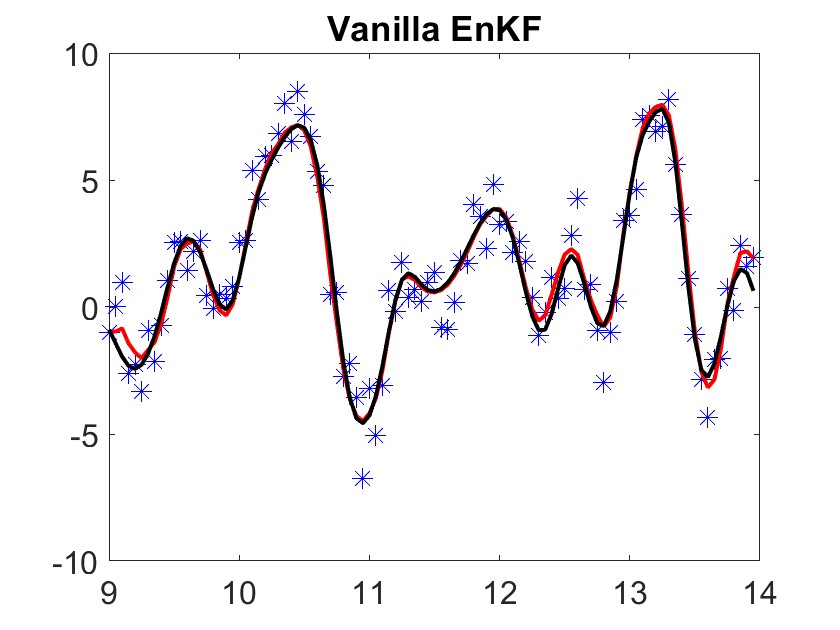} 
        \label{fig:subim1}
    \end{subfigure}
    \begin{subfigure}{0.32\textwidth}
        \includegraphics[width=\linewidth, height=6cm, keepaspectratio]{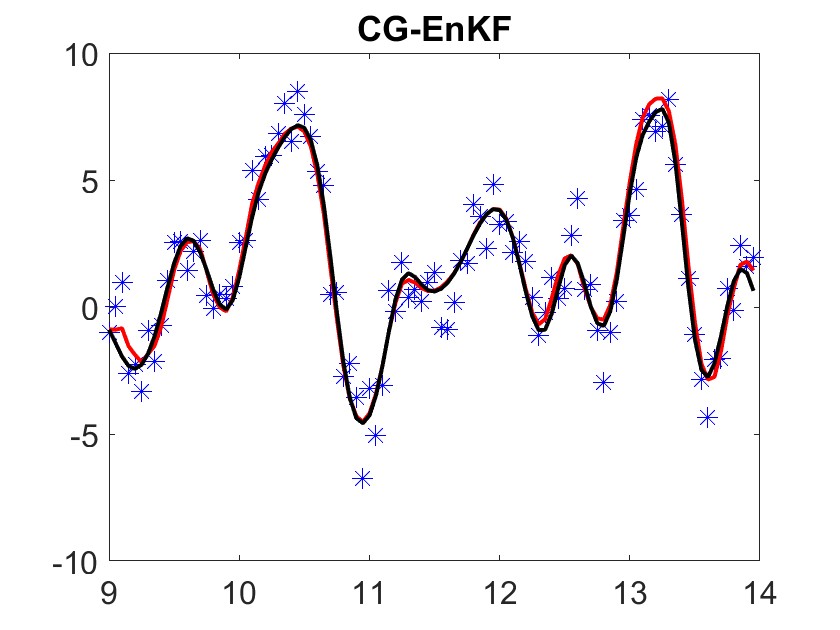}
        \label{fig:subim2}
    \end{subfigure}
    \begin{subfigure}{0.32\textwidth}
        \includegraphics[width=\linewidth, height=6cm, keepaspectratio]{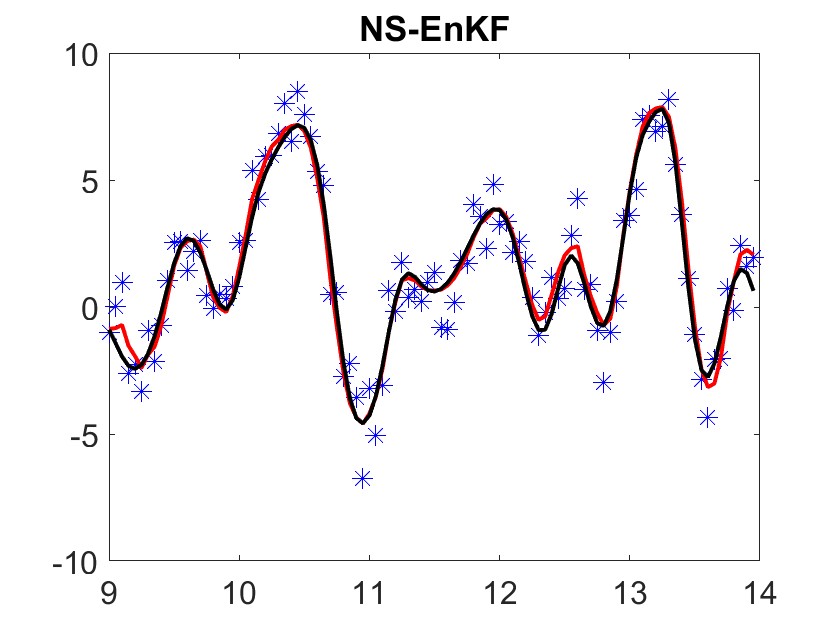}
        \label{fig:subim3}
    \end{subfigure}
    
    \begin{subfigure}{0.32\textwidth}
        \includegraphics[width=\linewidth, height=6cm, keepaspectratio]{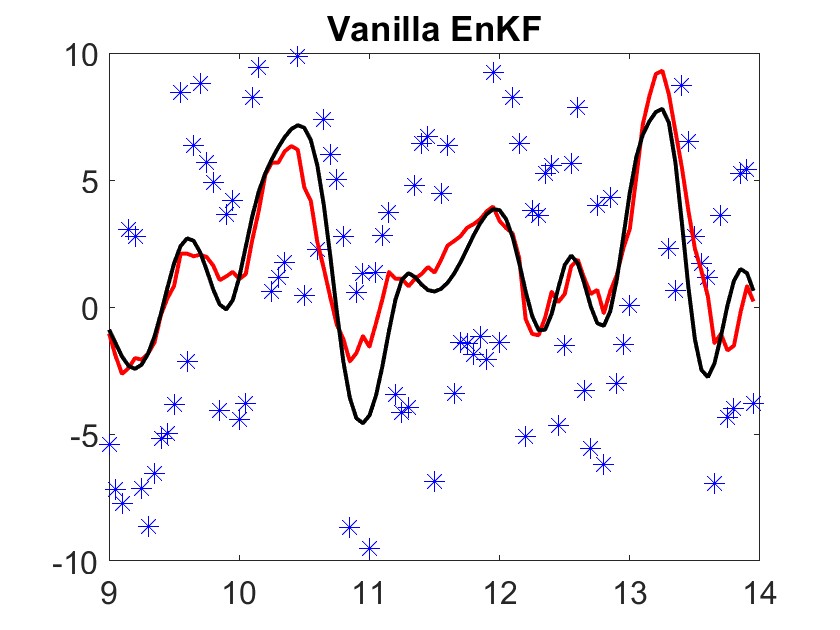} 
        \label{fig:subim4}
    \end{subfigure}
    \begin{subfigure}{0.32\textwidth}
        \includegraphics[width=\linewidth, height=6cm, keepaspectratio]{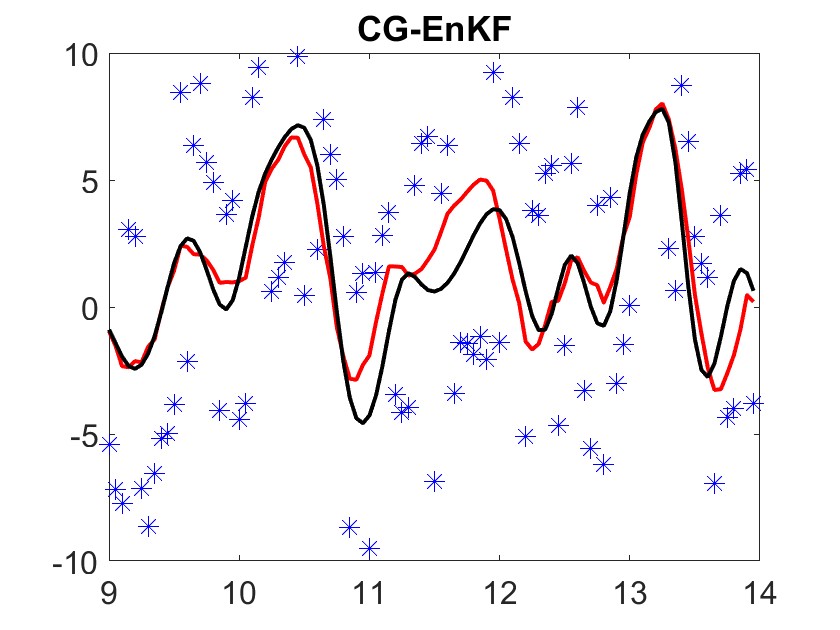}
        \label{fig:subim5}
    \end{subfigure}
    \begin{subfigure}{0.32\textwidth}
        \includegraphics[width=\linewidth, height=6cm, keepaspectratio]{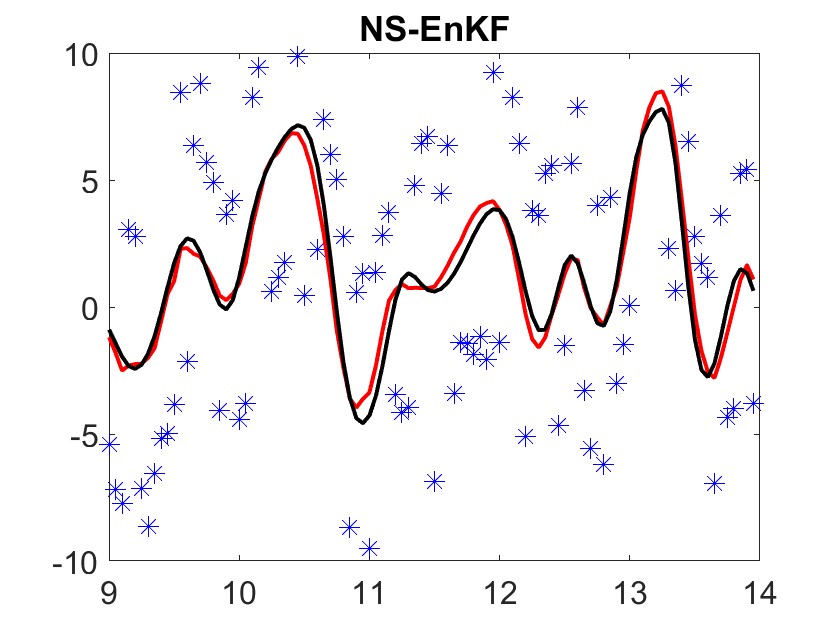}
        \label{fig:subim6}
    \end{subfigure}
    
    \begin{subfigure}{0.32\textwidth}
        \includegraphics[width=\linewidth, height=6cm, keepaspectratio]{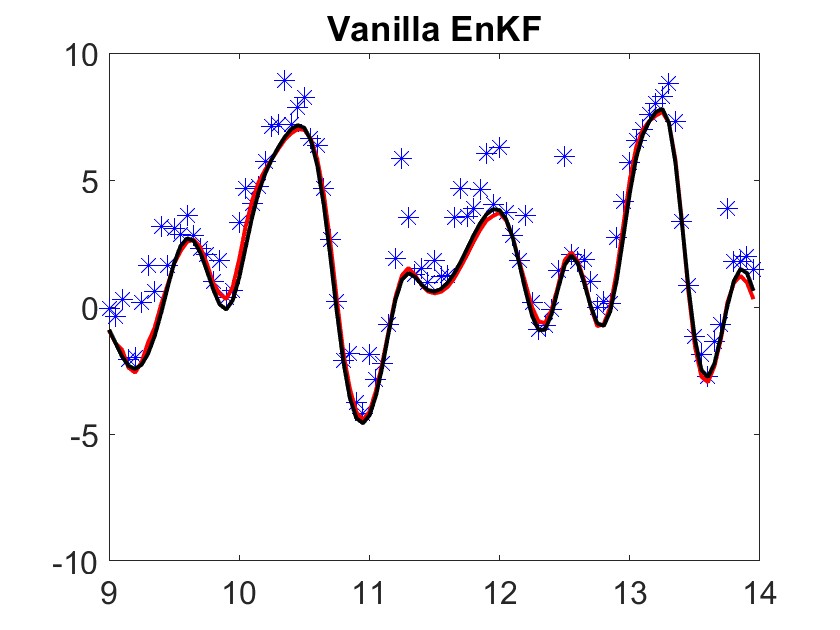} 
        \label{fig:subim7}
    \end{subfigure}
    \begin{subfigure}{0.32\textwidth}
        \includegraphics[width=\linewidth, height=6cm, keepaspectratio]{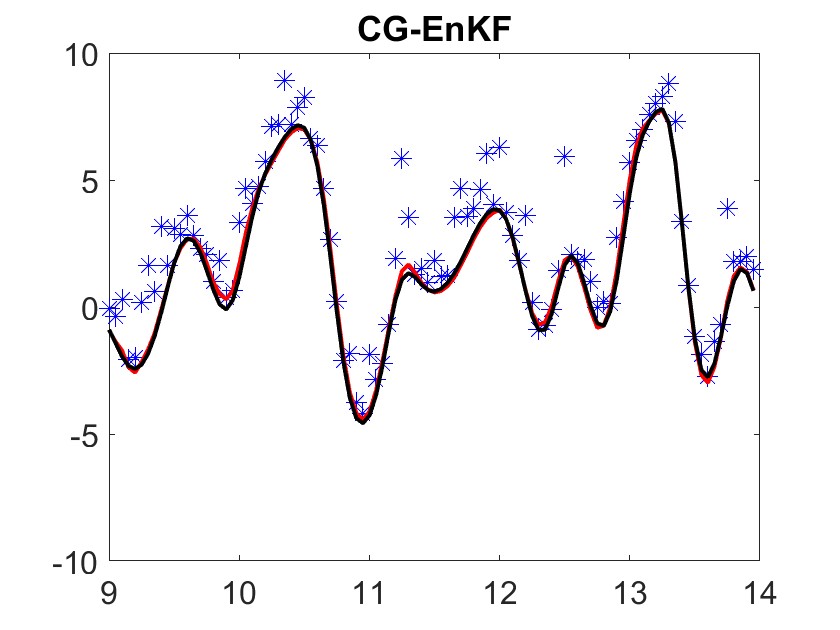}
        \label{fig:subim8}
    \end{subfigure}
    \begin{subfigure}{0.32\textwidth}
        \includegraphics[width=\linewidth, height=6cm, keepaspectratio]{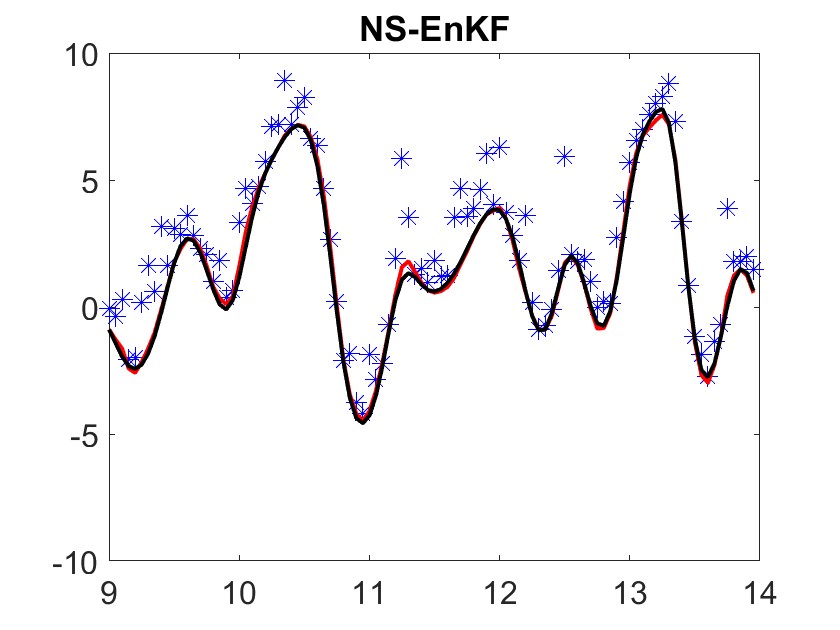}
        \label{fig:subim9}
    \end{subfigure}    
    \caption{Assimilated trajectories (red) against true trajectory (black) and cubic perturbed observations (blue stars) for vanilla EnKF (left), CG-EnKF (middle), and NS-EnKF (right), for the first dimension of L96 with Gaussian additive noise (top), bimodal additive noise (middle), and exponential additive noise (bottom).}
    \label{fig: L96 Dim 1 For Different Noises}
\end{figure}

In addition to linear and cubic observations as before, we also consider `Exponential' and `Bimodal', where the former simply adds exponentially distributed noise term with mean $1$ and the latter adds a bimodal distributed noise term, with modes at $-5$ and $5$. As before, all experiments were run on a single CPU.

The first $100$ steps of the first dimension of the assimilated trajectories are shown in Figure~\ref{fig: L96 Dim 1 For Different Noises}. Table~\ref{Table: EnKF Results for 5500 assimilation cycles} gives the summary statistics throughout the entire assimilation. We note that CG-EnKF and NS-EnKF perform very well across all noise types. Furthermore, the cost in wallclock time is reasonable.

We also tried to solve the filtering problem when the additive noise has Pareto distribution. These numbers were generated using MATLAB's \texttt{gprnd} with parameters $k=0.5$, $\texttt{sigma}=1$, and $\texttt{theta} = 2$ (following MATLAB's parameter naming convention). While our vanilla EnKF and CG-EnKF implementations blew up while trying to solve this problem, we did find that NS-EnKF could properly assimilate trajectories. The first 100 time steps of the first dimension of the assimilated trajectory is shown in Figure~\ref{fig: NSEnKF Pareto}. The summary statistics for this experiment are given in the last row of Table~\ref{Table: EnKF Results for 5500 assimilation cycles}.

Since SF cannot assimilate non-Gaussian additive noise (as per the discussion in the previous section), we did not include that model in our analysis for this set of experiments. We can also expect SF to take an inordinate amount of time to assimilate this problem within a reasonable range of accuracy. For instance, if $100$ assimilations took around $18$ minutes, as in Table~\ref{Table:Cubic SF vs EnKF Results}, then we may expect $5500$ assimilations to take around $1000$ minutes (or over $16$ hours). Clearly, SF more expensive than EnKF \textit{and} performed worse in our experiments.

\section{Conclusion}
This paper reviews two nonlinear extensions to the vanilla EnKF and compares them against a state of the art, machine learning based, algorithm called SF, for solving the Bayesian filtering problem. We perform data assimilation on a very simple system with nonlinear perturbations and find that the classic extensions, CG-EnKF and NS-EnKF, vastly outperform SF across multiple metrics such as accuracy and computational efficiency. Not only are CG-EnKF and NS-EnKF far cheaper to implement than SF, but they also beat SF in solving the filtering problem itself. Furthermore, SF as proposed by \cite{Bao24} cannot assimilate non-Gaussian observation perturbations: the likelihood function \eqref{Eq: SF Likelihood} explicitly requires this assumption (see also \cite[Section 2]{Bao24}. 

In contrast, CG-EnKF and NS-EnKF were able to assimilate non-Gaussian additive noise to a very satisfactory level of accuracy, as shown in Table~\ref{Table: EnKF Results for 5500 assimilation cycles}. Furthermore, these models were executed with a very reasonable computational cost, for example, CG-EnKF often assimilated trajectories in under three minutes. We believe CG-EnKF is a competitive approach to solve the Bayesian filtering problem and should be considered as a highly competitive baseline when developing novel (and potentially expensive) deep learning based methods for data assimilation.

\section*{Acknowledgements}

ZM acknowledges support from the NSF Mathematical Sciences Graduate Internship. RM acknowledges support from DOE Office of Science, Advanced Scientific Computing Research grant DE-FOA-0002493 (Data-intensive scientific machine learning) and the ARO Early Career Program award for Modeling of Complex Systems.

\bibliographystyle{siam}
\bibliography{bibliography.bib}

\end{document}